\documentclass[10pt,twocolumn,letterpaper，threeparttable]{article}

\usepackage{wacv}
\usepackage{times}
\usepackage{color}
\usepackage{epsfig}
\usepackage{graphicx}
\usepackage{amsmath}
\usepackage{amssymb}
\usepackage{subfigure}
\usepackage{adjustbox}
\usepackage{algorithm}
\usepackage{algpseudocode}
\usepackage{booktabs} 

\usepackage[pagebackref=true,breaklinks=true,colorlinks,bookmarks=false]{hyperref}

\wacvfinalcopy 

\def\wacvPaperID{961} 

\ifwacvfinal\fi
\begin{document}

\title{Supervised Learning Model for Key Frame Identification from Cow Teat Videos}

\author{Minghao Wang \hspace{2cm} Pinxue Lin \\
Yeshiva University\\
{\tt\small mwang3@mail.yu.edu}
}

\maketitle
\ifwacvfinal\thispagestyle{empty}\fi

\begin{abstract}

This paper proposes a method for improving the accuracy of mastitis risk assessment in cows using neural networks and video analysis. Mastitis, an infection of the udder tissue, is a critical health problem for cows and can be detected by examining the cow's teat. Traditionally, veterinarians assess the health of a cow's teat during the milking process, but this process is limited in time and can weaken the accuracy of the assessment. In commercial farms, cows are recorded by cameras when they are milked in the milking parlor. This paper uses a neural network to identify key frames in the recorded video where the cow's udder appears intact. These key frames allow veterinarians to have more flexible time to perform health assessments on the teat, increasing their efficiency and accuracy. However, there are challenges in using cow teat video for mastitis risk assessment, such as complex environments, changing cow positions and postures, and difficulty in identifying the udder from the video. To address these challenges, a fusion distance and an ensemble model are proposed to improve the performance (F-score) of identifying key frames from cow teat videos. The results show that these two approaches improve performance compared to using a single distance measure or model.

\end{abstract}

\section{Introduction}
Milk consumption holds significant importance in American society, and the United States proudly stands as the largest producer of milk worldwide. Renowned for its abundant nutritional content, milk is often considered a dietary necessity. The quality of milk is directly influenced by the health of the cows producing it. Among the various health issues affecting cows, mastitis stands out as one of the most critical problems. Detecting the risk of mastitis is crucial for ensuring the overall well-being and productivity of dairy cows. Traditionally, veterinarians have relied on assessing the health of a cow's teat during the milking process. However, this assessment is typically time-limited, spanning only tens of seconds, which poses limitations on the accuracy of mastitis risk evaluation.
In commercial dairy farms, cows are typically moved to the milking parlor where their milking sessions are recorded by cameras. Leveraging the advancements in computer vision and machine learning techniques, researchers have explored the utilization of recorded cow teat videos to identify key frames. Key frames refer to specific frames in the recorded videos where the cow's udder appears intact. By identifying and tagging these key frames, veterinarians gain the flexibility to conduct more comprehensive health assessments of the teat. This approach offers the potential to enhance the efficiency and accuracy of mastitis risk assessment in dairy cows.

Several challenges arise when utilizing cow teat videos for mastitis risk assessment. Firstly, the environment within the milking parlor can be complex, potentially resulting in issues such as insufficient lighting or obstructed views of the camera. These factors may adversely affect the quality and usability of the video data, impeding accurate analysis. Secondly, the position and posture of the cows in the videos can vary, and the shape of the milking machine can closely resemble that of the cow's teat. These similarities pose difficulties in distinguishing the udder from the milking machine, further complicating the identification process. Lastly, effective identification of cows' teats requires consideration of multiple factors, including teat size, shape, and color. The intricacies associated with these factors compound the challenge of accurately identifying key frames within the video footage.

To address the aforementioned challenges and improve the performance of mastitis risk assessment models, a fusion distance and an ensemble modeling techniques are proposed. The fusion distance method aims to enhance the model's ability to capture similarities between frames by integrating multiple distance metrics. By fusing these metrics, the model can better discriminate between key frames and non-key frames, ultimately improving the accuracy of the identification process. The ensemble model, on the other hand, leverages the strength of multiple individual models by combining their predictions. This ensemble approach harnesses the diverse perspectives and strengths of each model, resulting in an improved overall performance for mastitis risk assessment.

\section{Related Work}\label{sec:related}
The identification of key frames has long been a complex issue. Many traditional and neural network-based methods have been proposed.

\subsection{Traditional Video Key Frame Identification}
In the early days, most of the key frames were extracted based on the underlying features of the image, mainly including image color features, image texture features, image shape features, etc. The extraction methods for color features are usually based on the color histogram in RGB space, color histogram in HSV space, color aggregation vector, etc. The extraction methods for texture features are usually based on the LBP method, Markov random field model method, gray co-generation matrix, etc. The methods for shape features are usually based on the geometric parameter method, Fourier shape description method, wavelet descriptor, etc. Most of the existing feature extraction methods are based on the fusion of one or more features, but the underlying features of the image are usually extracted to a limited extent, and the advanced features of the image cannot be obtained\cite{ferraz2021comparison}.

The current method of key frame extraction is based on the traditional manual features of images, such as texture features, shape features, etc. However, this extraction method usually only extracts the relationship between adjacent frames of video frames, thus ignoring the dependence analysis of motion features before and after the frames that are farther away, and the phenomenon of missing frames occurs. Thus, some scholars use motion features to extract key frames, for example, by analyzing the optical flow field of video frames, and then extracting motion features based on the change of the motion field. Although this method improves the accuracy compared to color and other features, the extraction of optical flow field features is usually more complicated\cite{Paul2018}.

\subsection{Neural Network for Key Frame Identification}
Neural network models have a long history of development. In the last few decades, neural networks have moved from the laboratory into the industry. The development of neural network models began in 1943 and was inspired by the layout of biological neurons\cite{mcculloch1943logical}. These neural networks were built in layers, with the input of one layer connected to the output of the next. At that time mathematician Walter Pitts and neurophysiologist Warren McCulloch researched and wrote a paper describing how neurons might work and developed a simple neural network using electrical circuits\cite{piccinini2004first}.

By the 1980s, artificial neural networks (ANN) had become a hot topic in the field of artificial intelligence. Researchers realized that adding just a few hidden layers could greatly enhance the capabilities of their neural networks, leading to the development of ANN\cite{sejnowski2020unreasonable}. At the same time, advances in materials science, particularly the development of metal oxide semiconductor (MOS) very large scale integration (VLSI) in the form of complementary MOS (CMOS) technology, enhanced the performance of computers and enabled ANNs to move from theory to implementation.

Convolutional neural network (CNN) is a kind of neural network model commonly used in the field of computer vision, and its origin can be traced back to the "neocognitron" introduced by Kunihiko Fukushima in 1980\cite{Fukushima1980}. In the 21st century, CNN has developed rapidly. With advances in deep learning theory and improvements in numerical computing devices, CNN has made great progress in computer vision, natural language processing, and other fields\cite{Ferraz2021}. Deep learning, a subset of machine learning that uses neural networks with multiple hidden layers, has become the dominant approach in many fields, including computer vision, natural language processing, and speech recognition. By now, there have been many advances in the field of neural networks, including the development of new architectures, training algorithms, and applications. The use of neural networks has also spread beyond academia into industry, with many companies using them to develop new products and services.

Neural network models have developed rapidly in recent years. The field of computer vision has also made great progress with the introduction of several models. The ResNet (residual network) model was proposed by Kaiming He et al. in their paper “Deep Residual Learning for Image Recognition”\cite{He2015}. The ResNet model introduces the concept of residual learning, which is the use of fast connections to skip one or more layers. This allows the neural network to learn the residual function and thus improve the performance of image classification tasks. The Inception model was proposed by Christian Szegedy et al. in their paper “Going Deeper with Convolutions”\cite{Szegedy2015}. Inception introduces the concept of Inception modules, which are designed to efficiently extract features at multiple scales. The Xception model was proposed by François Chollet in his paper "Xception: Deep Learning with Depthwise Separable Convolutions" \cite{Chollet2017}. As the name implies, the Xception architecture performs slightly better than Inception V3 on the ImageNet dataset and significantly better than it on a larger image classification dataset. on a larger image classification dataset. The ViT (Vision Transformer) model was introduced by Alexey Dosovitskiy et al. in their paper “An Image is Worth 16x16 Words: Transformers for Image Recognition at Scale”\cite{Dosovitskiy2020}. The ViT model applies the transformer architecture originally designed for text-based tasks to image recognition. the ViT model represents the input image as a series of image patches and directly predicts the category labels of the image.

Neural network models have been used to extract video key frames. A key frame is usually a representative frame that contains all the facts or important information of a video collection. A key frame of a video is a still image that provides the most exact and compact summary of the video. It is also called a representative frame, R-frame, still-image abstract, or static storyboard. Key frames are used in several tasks like video browsing, video summarization, searching, understanding, and chapter titles in DVDs and are also used in many applications such as surveillance video and medical video. A combination model of CNN and Recurrent Neural Network (RNN) is to identify key frame\cite{Hu2018}. CNN is applied to estimate the pose probability of each frame from the selected foreground and extract key frames by the maximum probability of each pose. RNN can be used to process temporal information to improve the performance of video frames classification. Zhang et al. introduced an Unsupervised Few Shot Key Frame Extraction (UFSKE) model for cow teat videos\cite{Zhang2022}. They had also performed some other work to automate nipple health assessment using convolutional neural networks to classify the degree of nipple alteration on images\cite{zhang2022separable}. Wang et al. used the You Only Look Once v5 (YOLOv5) model to detect mastitis by thermal infrared images, based on the UST (difference in skin surface temperature) detection method between the left and right udder of cows\cite{Wang2022}.

\section{Methods}\label{sec:method}

\subsection{Motivation}
There are already videos that have been tagged with key frames \cite{Zhang2022}. Based on the uniqueness of the data and the problem, supervised learning is used to identify video key frames. 



\subsection{Pre-processing}
There is an upper limit to the memory of the device, and there is a certain compression of the information in the videos. Thus, for the given videos V as train sample, convert them to images $I=\left\{i_{0}+i_{1}\right\}$ on per-frame basis, where $i_0$ is ordinary frames and $i_1$ is key frames. This reduces the memory usage and reduces information loss.

\subsection{Fusion Distance}
3.3.1 Raw Deep Features

The deep features of video frames are extracted from models and obtained as the raw deep features. Deep features are extracted from the layer prior to the last fully connected layer. They are the high-level representations learned of video frames by the models.
\begin{equation}\label{eq:mse}
    d_{raw}=\operatorname{Model}(I),
\end{equation}
where $d_{raw}$ is the raw deep features, $\operatorname{Model}$ is the used neural network model, and the $I$ is video frames. The raw deep feature $d_{raw}$ is a vector of the shape (1, 2048).

3.3.2 Deep Distance

The task of cow teat identification from large video streams lacks a dedicated deep model that is specifically designed to tackle this particular challenge. As a result, alternative strategies need to be explored to address this gap in the literature. One promising approach is to leverage the power of pre-trained neural network models to extract deep features, which can serve as an effective method for cow teat identification in video frames. These pre-trained models are typically trained on extensive datasets for general visual recognition tasks, enabling them to learn rich and discriminative representations of visual information.

By extracting deep features from the pre-trained neural network models, denoted as $\Phi$, the comprehensive knowledge embedded within these models to enhance the performance of cow teat identification can be harness. These deep features encapsulate high-level visual patterns, abstract representations, and intricate characteristics relevant to the morphology and health of cow teats. The utilization of pre-trained models in this manner allows for the exploitation of the models' ability to capture complex visual cues and generalize well to novel data.

The extraction of deep features from pre-trained neural network models offers several advantages in the context of cow teat identification from video streams. Firstly, these models have undergone extensive training on large-scale datasets, enabling them to learn powerful and transferable representations. The deep features derived from such models possess a high level of information, allowing for a more comprehensive understanding of the visual content in the video frames. This facilitates the discrimination between key frames that exhibit signs of mastitis or other health issues and non-key frames that represent normal teat conditions.

The distance matrix is obtained by taking the absolute value of the difference between the original depth distance of the video frame minus the depth distance of the selected key frames. An element in a deep distance matrix is denoted as
\begin{equation}\label{eq:mse}
    M_{deep }=\left|\Phi_{(i)}-\Phi_{\left(i_{1}\right)}\right|,
\end{equation}
where $|\cdot|$ is absolute value of the difference between one video frame $\Phi_{(i)}$ and 32 randomly selected key frames $\Phi_{\left(i_{1}\right)}$. In this case, $M_{deep }$ is a matrix of the shape (32, 2048). Add each row of the deep distance matrix $M_{deep }$ to get a row vector of (1,2048). Then divide each element of this row vector by the number of rows of the deep distance matrix. The final result obtained is the deep distance. The deep distance is then defined as
\begin{equation}\label{eq:mse}
    d_{deep}=\operatorname{avg} M_{deep },
\end{equation}
where $\operatorname{avg}$ returns the average value of all rows in the matrix $M_{deep}$. This deep distance can represent feature differences between the current video frame to selected key frames. 

3.3.3 Fusion Distance

The raw distance and the deep distance are combined into a new distance function. Then Fusion distance can be got
\begin{equation}\label{eq:mse}
    D_{deep }=\left(d_{raw }, d_{deep }\right),
\end{equation}
where $D_{deep}$ is fusion distance. The fused deep distance $D_{deep}$ is a vector with (1, 4096). It has enriched information and dimensionality compared to the raw distance and the deep distance to improve the performance of models.

\subsection{Ensemble Model}
An ensemble model is a powerful machine learning technique that combines the predictions of multiple individual models to improve the overall performance. The fundamental idea behind ensemble methods is that by aggregating the predictions of several models, the resulting predictions will be more accurate and robust than those of any single model. This is because individual models may have different strengths and weaknesses, and by combining their predictions, the ensemble model can leverage their strengths while mitigating their weaknesses.

An ensemble model is created using five individual models. Each model determines whether the current frame is 0 or 1. When a model predicts that the current frame is 0, 0 gets one vote; when a model predicts that the current frame is 1, 1 gets one vote. Finally, if 0 has more votes than 1, the ensemble model predicts that the current frame is 0; otherwise, it predicts that the current frame is 1. This approach is known as majority voting and is a common technique used in ensemble models to combine the predictions of individual models. The ensemble model is represented as
\begin{equation}\label{eq:mse}
    i_{n}(0,1)=\sum V_{i}\left(i_{n}\right),
\end{equation}
where $V_{i}$ denotes the vote of the $i^{t h}$ models, $i_{n}$ is the $n^{t h}$ picture. By this method, more video frame features can be captured.

\subsection{Key Frame Identification Accuracy Evaluation}
The F-score is used to evaluate the accuracy of key frame identification. It, also known as the F1 score or F-measure, is a measure of a model's accuracy that takes into account both precision and recall\cite{vanRijsbergen1979}. Precision is the ratio of true positive predictions to the total number of positive predictions, while recall is the ratio of true positive predictions to the total number of actual positive instances.
\begin{equation}\label{eq:mse}
    Precision=\frac{2(\text { True Positives })}{(\text { True Positives }+ \text { False Positives })},
\end{equation}
\begin{equation}\label{eq:mse}
    Recall=\frac{2(\text { True Positives })}{(\text { True Positives }+ \text { False Negatives })},
\end{equation}
\begin{equation}\label{eq:mse}
    F=\frac{2( Precision  \times  Recall )}{( Precision +  Recall )},
\end{equation}

\subsection{Datasets}
Eighteen videos of the commercial farm milking process were shot with a tripod-mounted GoPro 10 camera and their key frames. These videos are in mp4 format and have a size of 2.57 GB, and the key frame information is stored in .mat files.

\begin{table}[h]
\caption{Cow teat videos}
\begin{tabular}{llll}
\#  & Video Name & Memory Size & Usage           \\
\hline
\hline
1   & GH010063   & 27.9MB      & Training Data   \\
2   & GH010065   & 116.0MB     & Training Data   \\
3   & GH010066   & 222.0MB     & Training Data   \\
4   & GH010067   & 24.2MB      & Training Data   \\
5   & GH010068   & 49.3MB      & Training Data   \\
6   & GH010069   & 37.7MB      & Training Data   \\
7   & GH010070   & 222.0MB     & Training Data   \\
8   & GH010071   & 217.0MB     & Training Data   \\
9   & GH010072   & 223.0MB     & Test Data       \\
10  & GH020066   & 222.0MB     & Test Data       \\
11  & GH020070   & 32.7MB      & Training Data   \\
12  & GH020071   & 169.0MB     & Training Data   \\
13  & GH020072   & 222.0MB     & Training Data   \\
14  & GH030066   & 221.0MB     & Training Data   \\
15  & GH030072   & 179.0MB     & Training Data   \\
16  & GH040066   & 197.0MB     & Validation Data \\
17  & GH050066   & 223.0MB     & Test Data       \\
18  & GH060066   & 22.3MB      & Validation Data \\
\hline
\hline
Ave &            & 145.93MB    &                
\end{tabular}
\end{table}

As shown in the table 1, there are 18 videos. Thirteen of them are training set, two are validation set, and three are test set. the average memory size of the 18 videos is 145.93 MB.

\subsection{Implementation Details}
Pre-trained ResNet-50, ResNet-101, Inception-V4, Xception, and ViT are used. The use of pre-trained neural network models and the extraction of deep features alleviate the need for training a specialized deep model from scratch, which can be computationally intensive and require substantial amounts of annotated data. By using pre-trained models, it is possible to speed up the training of models and obtain better results with relatively fewer computational resources.

Frame image features are extracted with an NVIDIA RTX 4090 GPU with 24 Gigabytes. Models are also run on this. The hyperparameters are set at the learning rate=0.00003, and weight decay=0.01. Since the data volume is large enough, a simple enhancement is performed before the video frames input models. Random horizontal flip probability is 0.5, resize is (224, 224), Normalize is set to mean=[0.485, 0.456, 0.406], std=[0.229, 0.224. 0.225]. The identification abilities of pre-trained ResNet50, ResNet101, Xception, Inception-V3, Inception-V4, and ViT for key frames are tested. The results are shown below. 

\begin{table}[h]
\caption{Performances of Pre-trained Models}
\begin{tabular}{llllll}
         & ResNet-50 & ResNet-101 & Inception   \\
\hline
\hline
GH020066 & 62.09     & 62.09      & 55.66       \\
GH010072 & 51.14     & 48.86      & 50.63       \\
GH050066 & 64.09     & 63.38      & 64.50       \\
\hline
\hline
Average  & 59.11     & 58.11      & 56.93
\end{tabular}
\end{table}
\begin{table}[h]
\centering
\begin{tabular}{llllll}
         & Xception & ViT   \\
\hline
\hline
GH020066 & 59.72    & 61.72 \\
GH010072 & 47.30    & 50.09 \\
GH050066 & 63.38    & 64.82 \\
\hline
\hline
Average  & 55.93    & 58.88
\end{tabular}
\end{table}

\subsection{Performance}
As the table below shows, the performances of the models after using the fusion distance and ensemble model are improved. It can be seen that after using the fusion distance, the ResNet model still performs the best on the dataset, while the worst-performing model on the dataset is still Xception.

\begin{table}[h]
\caption{Improved performance}
\begin{tabular}{lllllll}
         & ResNet-50 & ResNet-101 & Inception-V4\\
\hline
\hline
GH020066 & 63.72     & 63.51      & 64.81       \\
GH010072 & 54.45     & 52.53      & 53.15       \\
GH050066 & 64.92     & 65.60      & 66.05       \\
\hline
\hline
Average  & 61.03     & 60.54      & 61.33         
\end{tabular}
\end{table}
\begin{table}[h]
\centering
\begin{tabular}{lllllll}
         & Xception & ViT   & Ensemble Model \\
\hline
\hline
GH020066 & 60.51    & 61.72 & 66.76          \\
GH010072 & 50.50    & 51.82 & 54.58          \\
GH050066 & 61.19    & 64.76 & 67.59          \\
\hline
\hline
Average  & 57.40    & 59.43 & 62.97         
\end{tabular}
\end{table}

The performance of all models is improved after using the fusion distance. the Inception-V4 model has the largest improvement in the F-score of 4.4, and the ViT model has the smallest improvement in the F-score of 0.88. The average F-score improvement is 2.22. The identification of key frames is further improved after using the ensemble model. The F-score is improved by 5.178 relative to the initial pre-trained models and by 3.024 relative to the models with fusion distance.

\section{Discussion}\label{sec:dis}

While both the fusion distance and ensemble models have shown improvements in enhancing the identification of key frames in cow teat videos, the extent of the performance improvement remains limited. Several possible reasons can be attributed to this limited enhancement. First and foremost, the selection and tuning of model hyperparameters may not have been optimal. Due to constraints such as limited data availability and hardware limitations, the experiments might not have thoroughly explored different hyperparameter configurations. The effectiveness of the models can heavily depend on these hyperparameters, and fine-tuning them could potentially yield better results. Another factor contributing to the limited performance improvement could be the fusion distance approach itself. While the fusion distance method incorporates richer information and features from multiple distance metrics, it is possible that the integration of these metrics has inadvertently diminished the differentiation between key frames and non-key frames to some extent. Striking the right balance and determining the optimal combination of distance metrics is a complex task that requires careful exploration and analysis. Furthermore, the ensemble model's performance can be influenced by the varying performances of the individual models within the ensemble. Assigning appropriate weights to different models in the ensemble is crucial to effectively leverage their respective strengths. If the weights are not properly determined, it can lead to suboptimal performance and restrict the overall improvement achieved by the ensemble model.

Despite the limited improvement observed in the experimental results, it is important to note that the performance of the model did exhibit some degree of strengthening. This reaffirms the validity and potential efficacy of the approach presented in this paper. However, acknowledging the room for further enhancement, future work should focus on exploring alternative fusion distance and ensemble models that can potentially yield more substantial improvements. Several external factors may also contribute to the performance of this key frame identification method. The subject itself, in this case, the cows, and the environment in which the method is deployed can have an impact. Although efforts were made to minimize the influence of natural conditions on the farm and in the milking parlor, the artificial conditions present in the cows' environment may still affect the model's performance. Factors such as lighting conditions, obstructions in the camera view, and variations in teat status among different cows can introduce challenges and introduce additional variability to the key frame identification process. The presence of milkers attached to the cows' bodies, which may share similarities in shape with cow teats, can further complicate the identification process.

Lastly, the size and quality of the dataset used for training and evaluation can significantly affect the performance of the key frame recognition method. A larger and more diverse dataset may provide the model with a broader range of examples and improve its ability to generalize. Proper preprocessing techniques, such as data augmentation and cleaning, can also play a crucial role in enhancing the model's ability to recognize key frames accurately.

In summary, while the fusion distance and ensemble models have shown limited performance improvements in identifying key frames from cow teat videos, several factors such as suboptimal hyperparameter tuning, the fusion distance method's impact on differentiation, and appropriate model weighting in the ensemble can contribute to this limitation. Additionally, the subject and environmental factors, as well as the dataset size and quality, can influence the model's performance. Future research should explore alternative approaches, address these limitations, and strive for further advancements in cow teat key frame identification.

\section{Conclusion}\label{sec:conclusion}
In this paper, a novel method for identifying key frames of cow teats from videos is introduced. The proposed method involves combining the raw distance of each video frame with the deep distance between each video frame and a small number of keyframes to generate a fused distance. This fused distance contains richer information and features, which can improve the performance of the model in identifying key frames of cow teats. To further improve the performance of the model, an ensemble model that combines the predictions of multiple individual models is proposed. This ensemble model is voted by recording the results of every single model, and the result with the most votes is the output of this ensemble model. This approach leverages the strengths of multiple individual models while mitigating their weaknesses, resulting in more accurate and robust predictions. In addition, by converting video frames to images, the performance requirements of the training model on the device are reduced while releasing the compressed information in the video. The results show that these two proposed methods are able to improve the performance of the model in identifying key frames of cow teats, but the improvement is limited.



{\small
\bibliographystyle{unsrt}
\bibliography{egbib}
}

\end{document}